\documentclass[10pt,twocolumn,letterpaper,fleqn]{article}

\usepackage{tabularx} % also loads 'array' package
\newcolumntype{C}{>{\centering\arraybackslash}X} % centered version of 'X' columns
\usepackage{multirow}

\usepackage{rotating}

\usepackage[utf8]{inputenc}
\usepackage[T1]{fontenc}
\usepackage{textcomp}
\usepackage{booktabs}
\usepackage{stfloats}
\usepackage{iccv}
\usepackage{times}
\usepackage{epsfig}
\usepackage{graphicx}
\usepackage{amsmath}
\usepackage{amssymb}

\usepackage[utf8]{inputenc}
\usepackage{siunitx}
\graphicspath{ {/} }
\usepackage[caption=false]{subfig}
\usepackage{array}
\newcolumntype{P}[1]{>{\centering\arraybackslash}p{#1}}
\newcolumntype{M}[1]{>{\centering\arraybackslash}m{#1}}

\newcommand{\mybold}[1]{\boldsymbol{\mathbf{#1}}}

\newcommand{\bb}{\mybold{b}}
\newcommand{\cb}{\mybold{c}}
\newcommand{\fb}{\mybold{f}}

\newcommand{\hb}{\mybold{h}}
\newcommand{\ib}{\mybold{i}}
\newcommand{\ob}{\mybold{o}}

\newcommand{\vb}{\mybold{v}}
\newcommand{\wb}{\mybold{w}}
\newcommand{\xb}{\mybold{x}}
\newcommand{\yb}{\mybold{y}}
\newcommand{\zb}{\mybold{z}}

\newcommand{\Ab}{\mybold{A}}
\newcommand{\Fb}{\mybold{F}}
\newcommand{\Hb}{\mybold{H}}
\newcommand{\Ib}{\mybold{I}}
\newcommand{\Kb}{\mybold{K}}
\newcommand{\Qb}{\mybold{Q}}
\newcommand{\Rb}{\mybold{R}}
\newcommand{\Wb}{\mybold{W}}

\newcommand{\yest}{\hat{\mybold{y}}}
\newcommand{\zest}{\hat{\mybold{z}}}

\newcommand{\Pest}{\hat{\mybold{P}}}
\newcommand{\Qest}{\hat{\mybold{Q}}}
\newcommand{\Rest}{\hat{\mybold{R}}}
\newcommand{\Thetab}{\mybold{\theta}}

\newcommand*{\affaddr}[1]{#1} % No op here. Customize it for different styles.
\newcommand*{\affmark}[1][*]{\textsuperscript{#1}}
\newcommand*{\email}[1]{\texttt{#1}}

\usepackage[breaklinks=true,bookmarks=false]{hyperref}

\iccvfinalcopy % *** Uncomment this line for the final submission

\def\iccvPaperID{2544} % *** Enter the ICCV Paper ID here

\ificcvfinal\fi
\begin{document}

%%%%%%%%% TITLE
\title{Long Short-Term Memory Kalman Filters:\\Recurrent Neural Estimators for Pose Regularization}

\author{%
	Huseyin Coskun\affmark[1], Felix Achilles\affmark[2], Robert DiPietro\affmark[3], Nassir Navab\affmark[1,3], Federico Tombari\affmark[1]\\
	\affaddr{\affmark[1]Technische Universität München},
	\affaddr{\affmark[2]Ludwig-Maximilians-University of Munich},\\
	\affaddr{\affmark[3]Johns Hopkins University}\\
	\email{huseyin.coskun@tum.de, felix.achilles@med.lmu.de}\\
	\email{rdipietro@gmail.com, navab@cs.tum.edu, tombari@in.tum.de}
}
\maketitle
%\thispagestyle{empty}

%%%%%%%%% ABSTRACT
\begin{abstract}
One-shot pose estimation for tasks such as body joint localization, camera pose estimation, and object tracking are generally noisy, and temporal filters have been extensively used for regularization. One of the most widely-used methods is the Kalman filter, which is both extremely simple and general. However, Kalman filters require a motion model and measurement model to be specified a priori, which burdens the modeler and simultaneously demands that we use explicit models that are often only crude approximations of reality. For example, in the pose-estimation tasks mentioned above, it is common to use motion models that assume constant velocity or constant acceleration, and we believe that these simplified representations are severely inhibitive. In this work, we propose to instead learn rich, dynamic representations of the motion and noise models. In particular, we propose learning these models from data using long short-term memory, which allows representations that depend on all previous observations and all previous states. We evaluate our method using three of the most popular pose estimation tasks in computer vision, and in all cases we obtain state-of-the-art performance.
\end{abstract}

%%%%%%%%% BODY TEXT
\section{Introduction}
Pose estimation from images is a recurring challenge in computer vision, for example for tasks such as camera pose estimation, body joint localization, and object tracking. Such tasks have recently benefited from learned models \cite{Kendall2015, Tan2015, Bogo2016}, but various problems persist when applying one-shot pose estimation to video data.
In fact, disregarding temporal information can result in very noisy estimates and in the confusion of visually similar but spatially distinct image features, such as those that result from the left and right legs in the case of body joint localization.
For this reason, temporal filters are a popular approach for improving the accuracy of pose estimation.
Among these methods, because of their simplicity and general applicability, Kalman filters (KF)~\cite{kalman1960new} are an extremely widely-used choice. Moreover, the extended Kalman filter (EKF)~\cite{Welch2006} is capable of handling non linear systems for both the measurement and transition models.

\begin{figure}[t]
\centering
\includegraphics[width=.85\columnwidth]{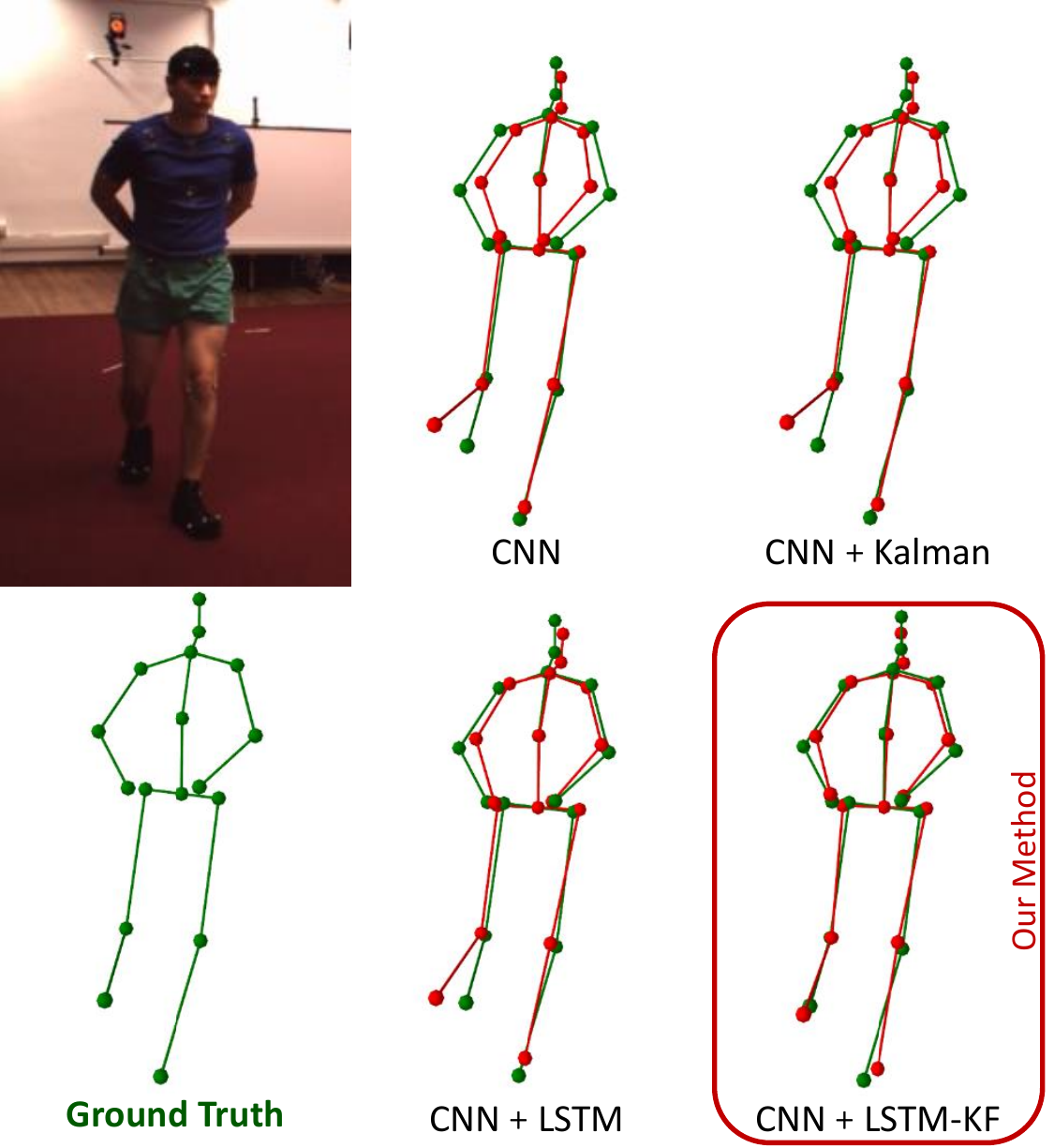}
\caption{The proposed LSTM-KF approach builds on Kalman filters and LSTM networks to yield an improved temporal regularizer for common pose estimation tasks such as 3D body landmark localization from RGB images.}
\label{fig:teaser}
\end{figure}

However, in many tasks, these measurement and transition models cannot be specified a priori, and in these situations the application of Kalman filters is severely limited. In particular, in these in these tasks we must devise carefully tuned measurement and transition models, and even once devised they tend to be overly simplistic. For example, in the aforementioned computer vision tasks the trajectories of objects and body parts do not follow any simple motion model. In such scenarios, Kalman filters are often applied under the assumptions of constant velocity or constant acceleration, which are clearly crude approximations to reality.

To overcome such limitations, attempts have been made to directly learn motion models from training data, for example with support vector machines (SVMs)~\cite{Salti2012} or with long short-term memory (LSTM) \cite{Krishnan2015}.
Learning motion models can alleviate the modeler from time-consuming Kalman filter selection and optimization and simultaneously enrich the underlying motion model.
However, using learned motion models to enforce temporal consistency in pose estimation has to cope with the constraint that sufficient training data needs to be available in order to cover all possible motion paths of the tracked object.

\begin{figure*}[t] 
\subfloat[LSTM-KF structure]{\includegraphics[width=1.0\columnwidth, height=3.7cm]{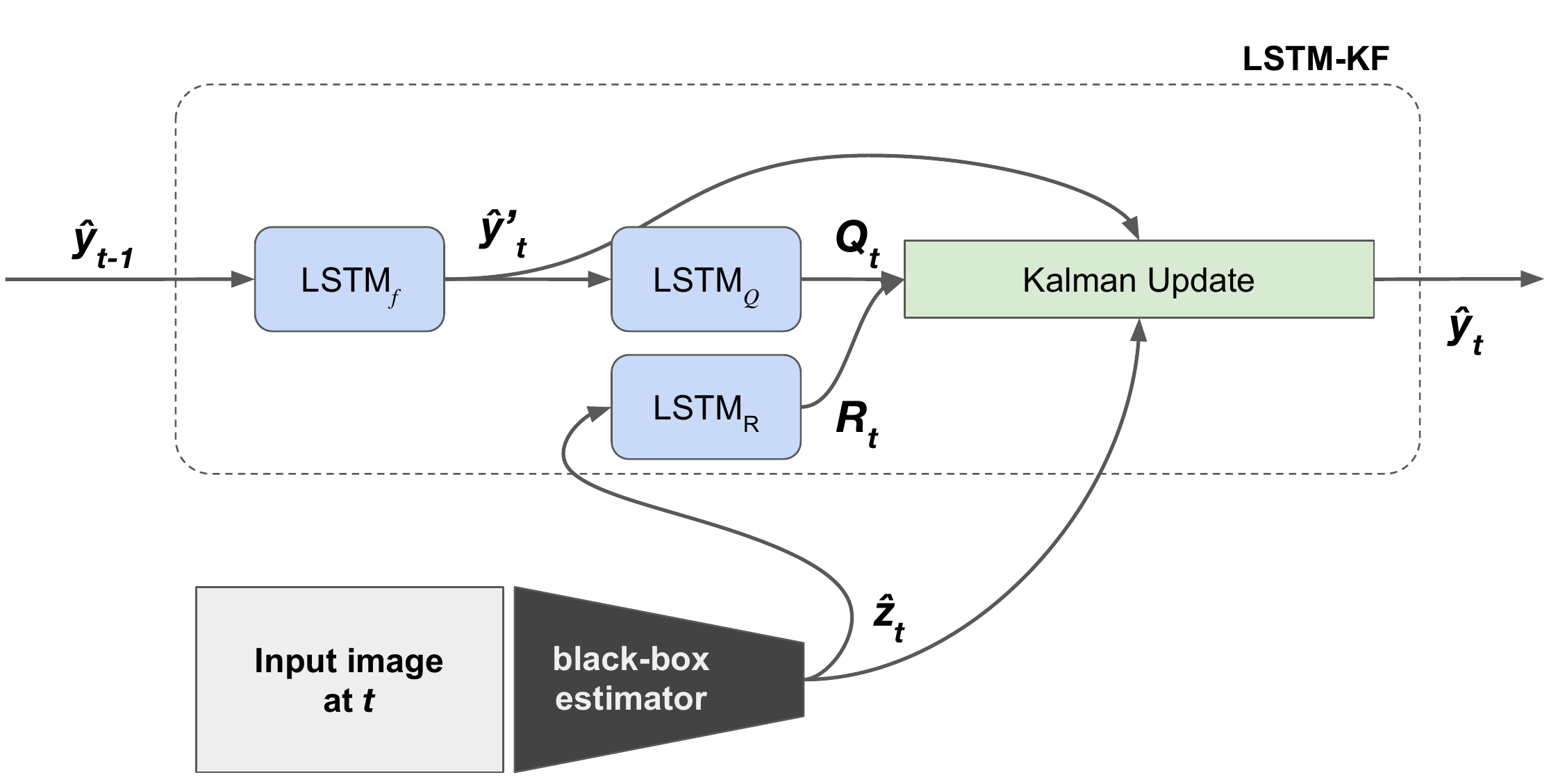}}
\subfloat[Unrolled LSTM-KF]{\includegraphics[width=1.1\columnwidth]{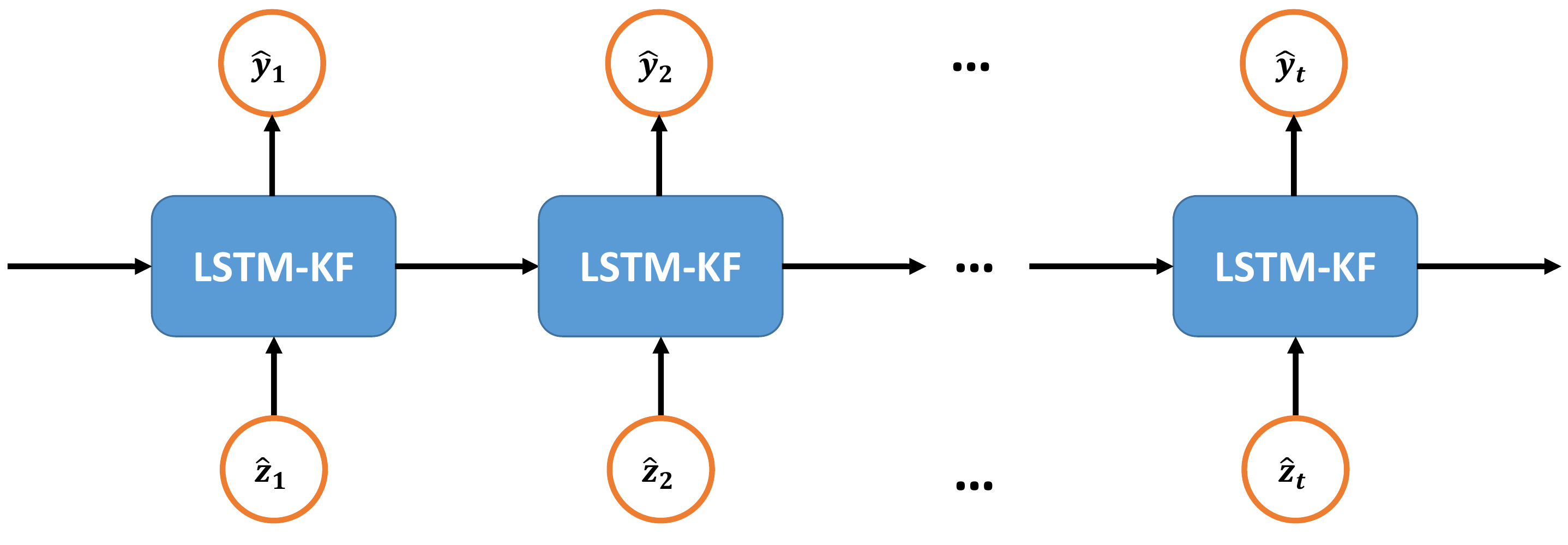}}
\caption{\textbf{Overview of the LSTM-KF.} (a) A high-level depiction of the architecture which uses three LSTM modules to predict the internals of the Kalman filter. (b) The LSTM-KF unrolled over time, which can be trained end to end with backpropagation through time.}
\label{fig:architecture}
\end{figure*}

In this work, we propose the LSTM Kalman filter (LSTM-KF), a new architecture which lets us learn the internals of the Kalman filter. In particular, we learn the motion model and all noise parameters of the Kalman filter, thus letting us gain the benefits of learning while letting us successfully train our models with less data.
The LSTM-KF architecture is illustrated in Fig.~\ref{fig:architecture}. This framework can be used to temporally regularize the output of any one-shot estimation technique, which from here forward will be considered a generic black-box estimator.

Specifically, our estimation model learns to predict the uncertainty of the initial prediction as well as the uncertainty of the incoming measurement, which is crucial in order to properly perform the update step.
In addition, a learned motion model is employed also for the prediction step.
Importantly, the estimator is not confined to the learned motion model, as it keeps on being refined by measurements during the update step.
As a result, the filter learns to implicitly regularize the pose over time without the need for a hand-crafted transition or measurement model.

We believe that our approach is advantageous with respect to learning-based Kalman filter techniques such as those in~\cite{Salti2012,Krishnan2015}. On one hand, in contrast to SVR \cite{Salti2012}, LSTM is able to estimate filter parameters using a model that depends on all previously observed inputs. On the other hand, by explicitly incorporating the prediction of LSTM with measurements in a Kalman update fashion, we relax the requirement on the LSTM to implicitly learn to fuse measurements with the state prediction for all possible motion paths, as attempted in~\cite{Krishnan2015}. 
Indeed, our model splits up the task of learning temporal regularization onto three distinct LSTMs that each have a defined objective: predicting the new state, estimating the prediction noise, and estimating the measurement noise.
Due to this split of objectives in a Kalman filter fashion, each individual LSTM learns a simpler task and our model will automatically start to rely on the measurements in case of low accuracy predictions. We evaluate the LSTM-KF using three relevant pose estimation tasks: body landmark localization, object tracking, and camera pose estimation, using real data from benchmark datasets. LSTM-KF outperforms both Kalman filters with different transition models and LSTM.

In the next section, we discuss related work. Next, we review Kalman filtering and long short-term memory in detail. In Section \ref{lstmkf}, we introduce the LSTM Kalman filter (LSTM-KF), including the underlying model, the modified prediction and update steps, and the full architecture which joins three LSTM modules with the Kalman filter. Next we move on to results, where we see LSTM-KF outperform other temporal regularization techniques, including standalone Kalman filters and standalone LSTM. Finally, we conclude and discuss future work.

\section{Related Work}
In recent literature, temporal regularization for pose estimation has been extensively studied.
We will first focus on those works that use an implicit regularization scheme and in the second part discuss those that explicitly use a learning-based Kalman filter architecture to infer temporal coherence.

For 3D human pose estimation, Du~\etal~\cite{Du2016} trained an overcomplete dictionary of body joint positions as well as joint velocities.
They use a Levenberg-Marquardt optimizer to find the dictionary basis coefficients that minimize the 2D backprojection error on the RGB input frame.
This way, joint velocities are used to regularize the joint position estimates.
In the experiments section we show that our approach yields superior results on the Human3.6M dataset.

Temporal regularization for 6 DOF object pose estimation was introduced by Krull~\etal~\cite{krull14}, who are using pose estimations from a random forest as input to a particle filter method.
The particle filter propagates a posterior distribution of the objects pose though time, using a predefined constant velocity motion model.
Choi~\etal extend the particle filter approach by introducing improved 3D features and a GPU implementation\cite{Choi2013}.

Two main lines of work can be identified that combine machine learning and Kalman filter models for temporal regularization.
We divide the approaches into those that learn static parameters of the Kalman filter and those that actively regress the parameters during filtering.
\textit{Static optimization} of noise covariance matrices was performed by Abbeel~et~al.~\cite{Abbeel2005}, who seek to replace manual fine-tuning of noise parameters in robotic navigation tasks.
The authors employ a coordinate ascent algorithm and optimize each individual element of the measurement and prediction noise covariance matrices.
However, this approach is only valid for noisy but time-invariant systems.
As opposed to our dynamic model, a change in measurement noise, for example due to partial occlusion of the tracked object, cannot be taken into account by their method and will therefore produce inaccurate state estimates.

Another approach is chosen by Krishnan~et~al.~\cite{Krishnan2015}, who focus on learning the underlying state transition function that controls the dynamics of a hidden process state.
However, only the state space equations of the Kalman filter are used, not the prediction and update scheme that performs optimally under the condition of linear state transitions and additive Gaussian noise~\cite{Welch2006}.
Instead, the authors train neural network models that jointly learn to propagate the state, incorporate measurement updates and react to control inputs.
Covariances were assumed to be constant throughout the estimation.
In our experiments section, we show that this approach produces inferior state estimations than a distinct prediction and update model, especially in the absence of large-scale training data.

\textit{Dynamic regression} of Kalman filter parameters was approached by Salti and Di~Stefano~\cite{Salti2012}.
In their work, support vector regression (SVR) is used to estimate a linear state transition function at each prediction step.
The prediction noise covariance matrix is estimated jointly with the transition function.
Their SVR based system is therefore able to deal with time-variant systems and outperforms manually tuned Kalman models on tracking tasks.
As opposed to our model, measurement noise covariances are kept constant.
The transition function is modeled as a matrix multiplication and can therefore only estimate linear motion models, while by design our model is able to estimate non-linear transition functions based on all previous state observations.

Haarnoja~et~al.~\cite{Haarnoja2016} focus on the integration of a one-shot estimation as measurement into a Kalman framework, but require the estimator to provide a prediction of the noise covariance together with the measurement.
The authors demonstrate a superior performance of their Kalman model by comparing to simple one-shot estimation and to a recurrent model that disregards measurement noise covariance.
In contrast, our model is designed to regard the estimator that provides measurement updates as a black-box system and automatically estimates the measurement noise covariance based on past observations, which enables us to combine it with existing one-shot estimators.

\section{Background}
In this section, we describe Kalman filters and long short-term memory (LSTM) and highlight the aspects of both methods which are most relevant to our LSTM Kalman filter, which we will describe in Section \ref{lstmkf}.

\subsection{Kalman Filters}
Kalman Filters (KFs) are optimal state estimators under the assumptions of linearity and Gaussian noise. More precisely, if we represent our state as $\yb_t$ and our measurement as $\zb_t$, and we assume the model
\begin{align}
\yb_t &= \Ab \yb_{t-1} + \wb, \quad \wb \sim N(\mybold{0}, \Qb) \label{kalmanmodel1}\\
\zb_t &= \Hb \yb_t + \vb, \quad \vb \sim N(\mybold{0}, \Rb) \label{kalmanmodel2}
\end{align}
where the matrices $\Ab$, $\Qb$, $\Hb$, and $\Rb$ are known, then the Kalman filter yields the best estimate $\yest_t$ in terms of sum-of-squares error.

The Kalman filter achieves optimality through an iterative feedback loop with two update steps, the prediction step and the update step. In the prediction step, we estimate the mean and covariance of our current state, independent of the current measurement:
\begin{align}
\yest'_t &= \Ab \yest_{t-1} \\
\Pest'_t &= \Ab \Pest_{t-1} \Ab^T + \Qb
\end{align}
In the update step, we compute the optimal Kalman gain $\Kb_t$ and use this along with our \emph{observed} measurement $\zest_t$ to estimate the mean and covariance of $\yb_t$:
\begin{align}
\Kb_t &= \Pest'_t \Hb^T (\Hb \Pest'_t \Hb^T + \Rb)^{-1} \\
\yest_t &= \yest'_t + \Kb_t (\zest_t - \Hb \yest'_t) \\
\Pest_t &= (\Ib - \Kb_t \Hb_t) \Pest'_t
\end{align}

\subsection{Long Short-Term Memory}
Recurrent neural networks (RNNs), unlike their feedforward counterparts, are naturally suited to modeling sequential data. However, early variants such as simple RNNs \cite{elman1990} were extremely difficult to train because of what is now known as the \emph{vanishing gradient problem} \cite{hochreiter1991, bengio1994}.

Long short-term memory (LSTM) \cite{hochreiter1997} was introduced specifically to address this problem, and has since become one of the most widely-used RNN architectures. In this work, we use the common variant with forget gates \cite{gers2000fg}, which are known to be crucial to achieving good performance \cite{greff2016}. This LSTM variant is described by
\begin{align}
\fb_t &= \sigma(\Wb_{fh} \hb_{t-1} + \Wb_{fx} \xb_t + \bb_f) \label{lstmbegin}\\
\ib_t &= \sigma(\Wb_{ih} \hb_{t-1} + \Wb_{ix} \xb_t + \bb_i) \\
\ob_t &= \sigma(\Wb_{oh} \hb_{t-1} + \Wb_{ox} \xb_t + \bb_o) \\
\tilde{\cb}_t &= \tanh(\Wb_{ch} \hb_{t-1} + \Wb_{cx} \xb_t + \bb_c) \label{lstmsimple} \\
\cb_t &= \fb_t \odot \cb_{t-1} + \ib_t \odot \tilde{\cb}_t \label{lstmupdate} \\
\hb_t &= \ob_t \odot \tanh(\cb_t) \label{lstmoutput}
\end{align}
where $\sigma(\cdot)$ denotes the element-wise sigmoid function and $\odot$ denotes element-wise multiplication. Focusing on Equations \ref{lstmupdate} and \ref{lstmoutput}, we can see that LSTM can be interpreted as resetting memory according to the forget gate $\fb_t$, writing to memory according to the input gate $\ib_t$, and reading from memory according to the output gate $\ob_t$, finally forming the output or \emph{hidden state}, $\hb_t$, at time step $t$. The intermediate memory cell $\tilde{\cb}_t$ and all gates depend on $\xb_t$, the input at the current time step, and on all $\Wb$ and $\bb$, which collectively form the parameters to be learned.

This architecture also easily extends to multiple-layer LSTM, where the hidden state $\hb_t$ from the first layer is simply treated as the input $\xb_t$ to the second layer,  or from the second to third layer, and so on.

\begin{figure*}[t]
\centering
 \includegraphics[width=.9\linewidth]{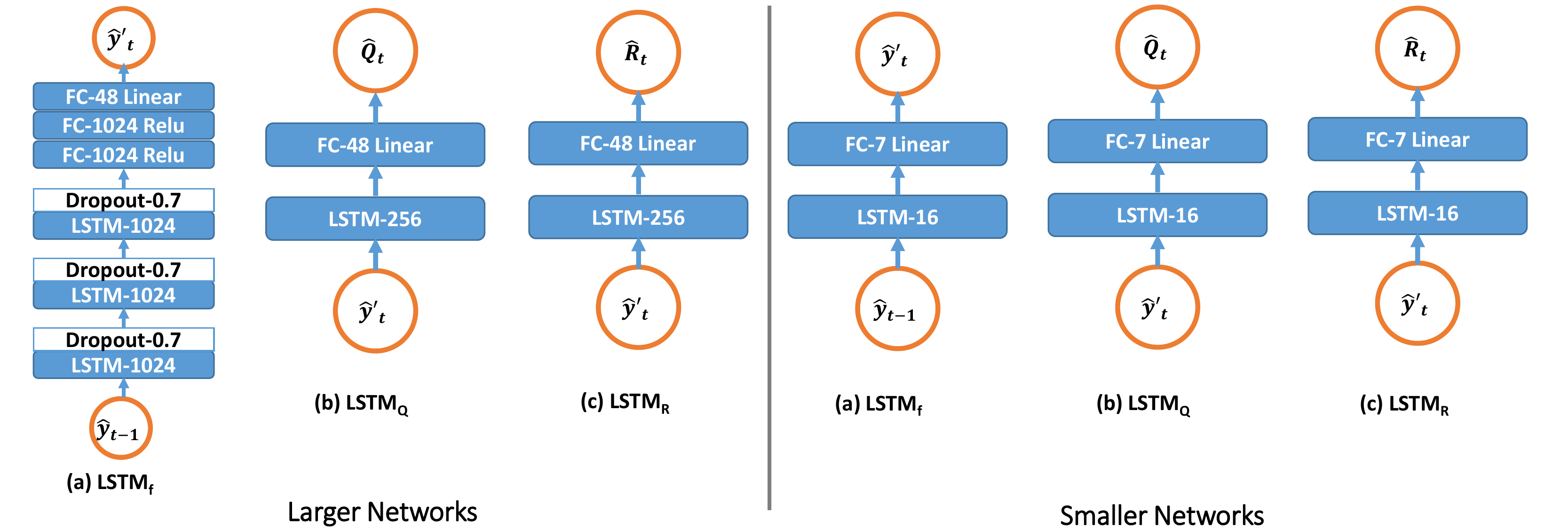}%
  \caption{\textbf{LSTM-KF architectures.} As detailed in Section \ref{experiments}, the larger networks are used for the \emph{Human 3.6M} dataset, and the smaller networks for all other (smaller) datasets.}
  \label{architecturedetails}
\end{figure*} 

\section{LSTM Kalman Filters}
\label{lstmkf}
In this section, we present the long short-term memory Kalman filter (LSTM-KF), a model for the temporal regularization of pose estimators. The main idea is to leverage Kalman filters \emph{without} the need to specify a linear transition function $\Ab$ or fixed process and measurement covariance matrices $\Qb$ and $\Rb$. Instead, we will model a nonlinear transition function $f$ along with $\Qb$, and $\Rb$ using three different long short-term memory (LSTM) networks, thus providing our model with the ability to learn rich, dynamic Kalman components from data.

\subsection{Model}
We always assume that incoming measurements are noisy estimates of the underlying state, and thus $\Hb = \Ib$ in Equation \ref{kalmanmodel2}. Equations \ref{kalmanmodel1} and \ref{kalmanmodel2} then take on the modified form
\begin{align}
\yb_t &= f(\yb_{t-1}) + \wb_t, \quad \wb_t \sim N(\mybold{0}, \Qb_t) \label{lstmkfmodel1}\\
\zb_t &= \yb_t + \vb_t, \quad \vb_t \sim N(\mybold{0}, \Rb_t) \label{lstmkfmodel2}
\end{align}
which specifies the underlying model of the LSTM-KF.

\subsection{Prediction and Update Steps}
Our prediction step is then defined by
\begin{align}
\yest'_t &= f(\yest_{t-1}) \\
\Pest'_t &= \Fb \Pest_{t-1} \Fb^T + \Qest_t \label{Pestprime}
\end{align}
where $f$ is modeled by one LSTM module, $\Fb$ is the Jacobian of $f$ with respect to $\yest_{t-1}$, and $\Qest_t$ is the output of a second LSTM module. Finally, our update step is
\begin{align}
\Kb_t &= \Pest'_t (\Pest'_t + \Rest_t)^{-1} \\
\yest_t &= \yest'_t + \Kb_t (\zest_t - \yest'_t) \\
\Pest_t &= (\Ib - \Kb_t) \Pest'_t \label{Pest}
\end{align}
where $\Rest_t$ is the output of a third LSTM module and where $\zest_t$ is our observed measurement at time $t$. Next we describe these LSTM modules in detail.

\subsection{Architecture}
We denote the three LSTM modules for $f$, $\Qest_t$, and $\Rest_t$ by LSTM$_f$, LSTM$_Q$, and LSTM$_R$; each is depicted in Fig. \ref{architecturedetails}, and an overview of the system is depicted in Fig. \ref{fig:architecture}.

At each time step $t$, LSTM$_f$ takes in the previous prediction $\yest_{t-1}$ as input and produces the intermediate state $\yest'_t$ (which does not depend on the current measurement). LSTM$_Q$ then takes $\yest'_t$ as input and produces an estimate of the process covariance, $\Qest_t$, as output. Meanwhile, the observation $\zb_t$ serves as input to LSTM$_R$, which only produces an estimate of the measurement covariance, $\Rest_t$, as output. Finally, $\yest'_t$ and $\zb_t$, along with our covariance estimates, are fed to a standard Kalman filter, as described by Equations \ref{Pestprime} through \ref{Pest}, finally producing the new prediction $\yest_{t}$.

We remark that in this work $\Qb$ and $\Rb$ are restricted to be diagonal, and they are restricted to be positive definite by exponentiating the outputs of the LSTM$_Q$ and LSTM$_R$ modules.

\subsection{Loss}
In preliminary experiments, we used standard Euclidean loss summed over all time steps, but in this case we found that the LSTM$_f$ module would fail to learn any reasonable mapping. Because of this, we added a term to our loss to enhance gradient flow to the LSTM$_f$ block, resulting in the loss
\begin{equation}
\label{loss}
L(\Thetab)= \frac{1}{T} \sum_{t=1}^T \lVert \yb_{t} - \yest_{t}(\Thetab) \rVert^2 + \lambda \lVert \yb_{t} - \yest'_{t}(\Thetab) \rVert^2
\end{equation}
We set the hyperparameter $\lambda$ to 0.8 using the Human3.6M dataset and kept it fixed for all other experiments, as we found that performance was relatively insensitive around this value.

\subsection{Optimization}
Our objective is to optimize all parameters $\Thetab$ to minimize the loss given by Equation \ref{loss} with respect to all free parameters in our model, which are a concatenation of all weight matrices and biases from all three LSTM modules. (Note that these modules are combinations of LSTM layers and linear layers, as depicted by figure \ref{architecturedetails}.)

Our model can be trained end to end, with gradients obtained using the backpropagation through time algorithm\cite{werbos1990}, which we implement using the TensorFlow framework \cite{abadi2016}. We use gradient updates according to the Adam \cite{kingma2014} optimizer.
\begin{figure}[t]
\centering
\includegraphics[width=0.4\textwidth]{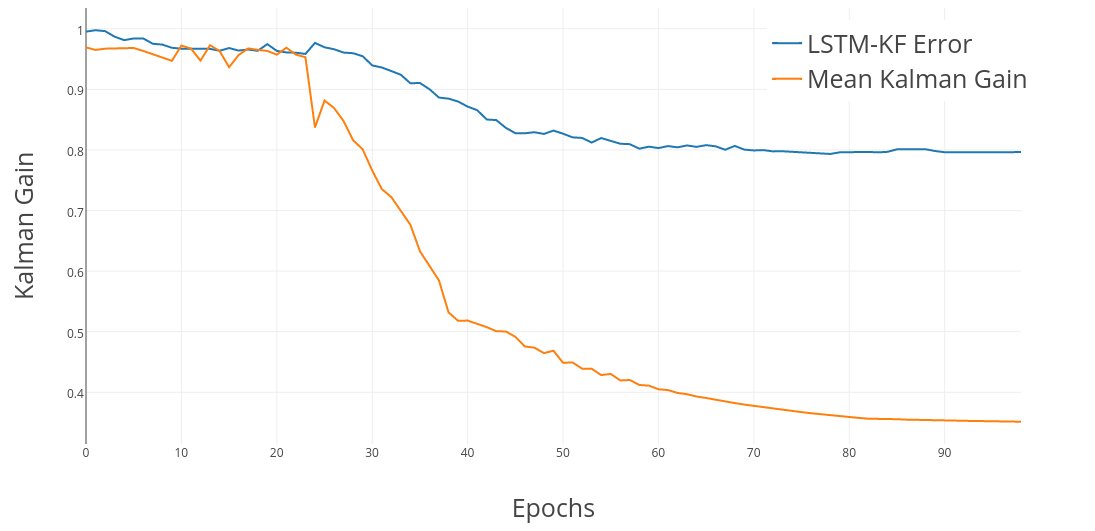}
\caption{\textbf{LSTM-KF error and mean Kalman gain during training.} At the beginning of training, the Kalman gain (as well as error) is high, indicating that the model is relying almost entirely on measurements. As training progresses, the Kalman gain drops considerably, indicating that the Kalman filter relies significantly on both on the measurements and the LSTM$_f$ module's output.}
\label{fig:ktime}
\end{figure}

\section{Experiments}
\label{experiments}
In this section we compare the pose estimation performance of our LSTM-KF architecture to a range of temporal regularization methods, including two standard Kalman filters that assume either a constant velocity or constant acceleration motion (respectively \textit{Kalman Vel, Kalman Acc}), to an exponential moving average filter (\textit{EMA}), and to a standard LSTM module (\textit{Std. LSTM}).
Specifically, this LSTM model that we compare to is a representative of the class of models proposed in~\cite{Krishnan2015}, and it is characterized by implicitly learning the prediction step as well as the measurement update step in an end-to-end fashion.

We evaluate these models on four different datasets, one for 3D human pose estimation, two for camera pose estimation, and one for object pose estimation, all of them using RGB images as input modality~\cite{Ionescu2014,Kendall2015,shotton2013scene}.

\subsection{Implementation Details}
We initialize all LSTM state-to-state weight matrices as random orthogonal matrices, all other LSTM weight matrices using a uniform distribution over $[-0.01, 0.01]$, and all linear-layer weight matrices using Xavier initialiation \cite{glorot2010}. All biases are initialized with zeros except for LSTM forget-gate bias; following best practices, we set these biases to 1.0 \cite{gers2000fg, greff2016}.

Noise covariance matrices of the Kalman filter methods (\textit{Kalman Vel, Kalman Acc}) as well as the window size of the exponential moving average method (\textit{EMA}) were optimized via grid search.

\subsection{Human Pose Estimation} 
\begin{table*}[t]
\centering
\begin{tabular}{l c c c c c c c c} 
\hline
& Directions & Discussion & Eating & Greeting & Phoning & Photo & Posing & Purchases\\
\hline
Li \etal \cite{li2015} & - & 136.88 & 96.94 & 124.74 & - & 168.68 & - & -\\
Tekin \etal \cite{tekin2016} & 102.39 & 158.52 & 87.95 & 126.83 & 118.37 & 185.02 & 114.69 & 107.61\\
Zhou \etal \cite{zhou2016} & 87.36 & 109.31 & 87.05 & 103.16 & 116.18 & 143.32 & 106.88 & 99.78\\
SMPLify \cite{Bogo2016} & 62.0 & \textbf{60.2} & 67.8 & 76.5 & 92.1 & \textbf{77.0} & 73.0 & 75.3\\
\hline
Inception & 67.18 & 74.79 & 71.80 & 73.85 & 81.04 & 88.73 & 72.58 & 73.12\\
+ {\small Kalman Vel.} & 67.70 & 74.01 & 71.73 & 73.32 & 80.74 & 88.03 & 72.22 & 73.45\\
+ {\small Kalman Acc.} & 67.08 & 74.75 & 71.21 & 73.23 & 80.74 & 88.01 & 72.11 & 73.31\\
+ {\small EMA} & 67.01 & 74.78 & 71.81 & 73.81 & 81.04 & 88.70 & 72.50 & 72.02\\
+ {\small Std. LSTM} & 62.70& 70.11 & 63.53 & 67.24 & 75.42 & 85.37 & 67.42 & 67.07\\
+ {\small LSTM-KF (\bf{ours})} & \textbf{61.41} & 69.98 & \textbf{62.12} & \textbf{65.93} & \textbf{71.93} & 83.92 & \textbf{63.0} & \textbf{65.87}\\
\hline \hline
& Sitting & SitDown & Smoking & Waiting & WalkDog & Walk & WalkTogether & \bf{Mean}\\
\hline
Li \etal \cite{li2015} & - & - & - & - & 132.17 & 69.97 & - & -\\
Tekin \etal \cite{tekin2016} & 136.15 & 205.65 & 118.21 & 146.66 & 128.11 & 65.86 & 77.21 & 125.28\\
Zhou \etal \cite{zhou2016} & 124.52 & 199.23 & 107.42 & 118.09 & 114.23 & 79.39 & 97.70 & 113.01\\
SMPLify \cite{Bogo2016} & 100.3 & 137.3 & 83.4 & 83.4 & 79.7 & 86.8 & 81.7 & 82.3\\
\hline
Inception & 91.36 & 111.19 & 79.25 & 71.67 & 88.04 & 71.95 & 74.01 & 79.8\\
+ {\small Kalman Vel.} & 91.04 & 111.1 & 79.01 & 71.90 & 87.99 & 87.99 & 74.35 & 79.20\\
+ {\small Kalman Acc.} & 90.88 & 111.11 & 79.13 & 71.51 & 87.62 & 87.62 & 74.10 & 79.07\\
+ {\small EMA} & 91.31 & 111.11 & 79.21 & 71.70 & 88.04 & 71.91 & 73.97 & 79.26\\
+ {\small Std. LSTM} & 85.15 & 104.16 & 72.69 & 72.68 & 80.77 & 59.23 & 61.36 & 73.22\\
+ {\small LSTM-KF (\bf{ours})} & \textbf{84.81} & \textbf{98.85} & \textbf{69.79} & \textbf{65.88} & \textbf{79.44} & \textbf{55.32} & \textbf{60.29} & \textbf{70.98}\\
\hline
\end{tabular}
\caption{Average 3D joint error on \emph{Human 3.6M} for test subjects 9 and 11. The error is given in [mm]. \label{tbl:human36m}} 
\end{table*}

The \textit{Human3.6M} dataset of Ionescu~\etal~\cite{Ionescu2014}, consists of 3.6 million RGB video frames from video sequences that were recorded in a controlled indoor motion capture setting.
In each of these sequences, one out of seven actors performs 15 activities with varying levels of movement complexity.
Each of the activities is between 3,000 and 5,000 frames long.
In our experiments, we follow the same data partition scheme as \cite{Bogo2016,zhou2016} for training and test set: training has 5 subjects (\textit{S1, S5, S6, S7, S8}) and test data 2 subjects (\textit{S9, S11}).
Similar to \cite{Bogo2016} we compute the model performance in terms of average Euclidean distance between estimated and ground-truth 3D joint positions.
Furthermore, following previous works for this dataset, we express all joint positions relative to a root joint, which is the pelvis joint in our case.
In order to get initial 3D human pose estimations on the RGB videos, we refine a Inception-v4 CNN model that was pre-trained on ImageNet~\cite{szegedy2016inception}.
For this fine tuning, we use a batch size of 30 and set the initial learning rate to 0.01 and reduce it about a decay factor of 10 at each epoch, and train for a total of only 3 epochs.
To prevent overfitting, we augment the RGB data by randomly cropping $300\times300$ patches from the $350\times350$ input images and randomly distort the brightness, hue, saturation and contrast of each input image.
Besides data augmentation, we apply dropout in the last layer, retaining values with a probability of $0.8$.
Re-training the network for the pose estimation task on a Tesla K40 GPU took 10 days.
We then use the Inception-v4 estimation values as measurement inputs to train the LSTM-KF and standard LSTM model.
\begin{figure}[t]
	\centering
	\includegraphics[width=1\linewidth]{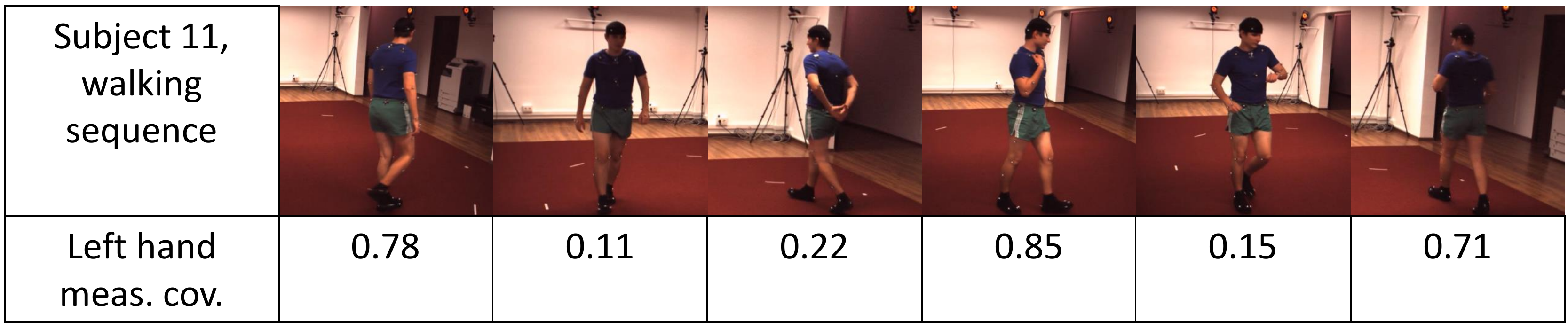}%
	\caption{\textbf{Measurement noise covariance during occlusion.} Here we include the Euclidean norm of covariance coefficients for the left hand (normalized between 0 and 1) along with the corresponding images from a \textit{Walking} test sequence. The model has learned to assign high measurement uncertainty to those frames in which the left hand is occluded.}
	\label{co_image}
\end{figure} 

\begin{table*}[t]
\setlength\tabcolsep{2 pt}
\centering
\begin{tabular}{l c c c c c c c c c c c c c c c c}  
\hline
 & \multicolumn{2}{c}{Chess}  &  \multicolumn{2}{c}{Fire} & \multicolumn{2}{c}{Heads}& \multicolumn{2}{c}{Office}& \multicolumn{2}{c}{Pumpkin}& \multicolumn{2}{c}{R. Kitchen}& \multicolumn{2}{c}{Stairs}& \multicolumn{2}{c}{\bf{Mean}}\\
 & \emph{tran.} & \emph{rot.} & \emph{tran.} & \emph{rot.} & \emph{tran.} & \emph{rot.} & \emph{tran.} & \emph{rot.} & \emph{tran.} & \emph{rot.} & \emph{tran.} & \emph{rot.} & \emph{tran.} & \emph{rot.} & \emph{tran.} & \emph{rot.} \\
 \hline
 PoseNet \cite{Kendall2015} & 0.38 & 7.51°& 0.47 & 16.61° & 0.32 & 13.6° & 0.48 & 7.79°& 0.54 & 11.17°& 0.59 & 9.14° & 0.55 & 15.65° & 0.50 & 11.47°\\
 + {\small Kalman Vel.} & 0.38 & 8.35°& 0.47 & 16.66° & 0.32 & 14.73° & 0.48 & 8.64°& 0.54 & 12.06°& 0.59 & 9.94° & 0.54 & 16.58° & 0.50 & 12.40°\\
 + {\small Kalman Acc.} & 0.37 & 8.34°& 0.47 & 16.67° & 0.32 & 14.71° & 0.48 & 8.62°& 0.54 & 12.09°& 0.59 & 9.95° & 0.54 & 16.58° & 0.49 & 12.39°\\
 + {\small EMA} & 0.37 & 7.31°& 0.47 & 16.46° & 0.32 & 13.53° & 0.47 & \textbf{7.48°}& 0.54 & 11.01°& 0.53 & 8.85° & 0.55 & 15.56° & 0.49 & 11.29°\\
 + {\small Std. LSTM} & 0.41 & 8.4°& 0.5 & 17° & 0.35 & 15.05° & 0.48 & 9.99°& 0.53 & \textbf{10.38°}& \textbf{0.51} & 9.71° & 0.65 & \textbf{13.62°} & 0.51 & 11.75°\\
 + {\small LSTM-KF (\bf{ours})}  & \textbf{0.33} & \textbf{6.9°}& \textbf{0.41} & \textbf{15.7°} & \textbf{0.28} & \textbf{13.01°} & \textbf{0.43} & 7.65°& \textbf{0.49} & 10.63°& 0.57 & \textbf{8.53°}& \textbf{0.46} & 14.56° & \textbf{0.44} & \textbf{10.83°}\\
\hline
\end{tabular}
\caption{Comparison of temporal regularisation methods on camera pose estimations provided by PoseNet on the \emph{7 Scenes} dataset. As in \cite{Kendall2015}, values are given as median errors in translation [m] and rotation [degrees].} 
\label{tbl:sevenscenes}
\end{table*}

\begin{table*}[t]
\setlength\tabcolsep{6 pt}
\centering
\begin{tabular}{l c c c c c c c c c c c c}  
\hline
& \multicolumn{2}{c}{Street}  &  \multicolumn{2}{c}{K. College} & \multicolumn{2}{c}{S. Facade}& \multicolumn{2}{c}{St. M. Church}& \multicolumn{2}{c}{Old Hospital}& \multicolumn{2}{c}{\bf{Mean}}\\
 & \emph{tran.} & \emph{rot.} & \emph{tran.} & \emph{rot.} & \emph{tran.} & \emph{rot.} & \emph{tran.} & \emph{rot.} & \emph{tran.} & \emph{rot.} &\emph{tran.} & \emph{rot.} \\
\hline
  PoseNet \cite{Kendall2015} & 3.35 & 6.12°& 1.97 & 5.38° & 1.65 & 8.49° & 2.88 & 9.04°& 2.60 & 5.32°& 2.49 & 6.87  \\
 + {\small Kalman Vel.} & 3.16 & 5.93°& \textbf{1.85} & 5.29° & \textbf{1.48} & 8.20° & 2.94 & 9.29°& 2.53 & 5.07°& 2.39& 6.75°\\
 + {\small Kalman Acc.} & 3.14 & 5.92°& 1.88 & 5.29° & 1.49 & 8.33° & 2.95 & 9.33°& 2.45 & 5.07°& 2.38& 6.79°\\
 + {\small EMA} & 3.33 & 5.63 °& 1.95 & \textbf{5.28°} & 1.62 & 8.35° & 2.82 & 8.99°& 2.68 & 5.10°& 2.48& 6.67°\\
 + {\small Std. LSTM} & 9.56 & 11.2°& 4.24& 7.95° & 1.87 & 7.04° & 3.34 & 11.52°& 4.03 & 6.46° & 4.61& 8.83°\\
 + {\small LSTM-KF (\bf{ours})} & \textbf{3.05} & \textbf{5.62°}& 2.01 & 5.35° & 1.63 & \textbf{6.89°} & \textbf{2.61} & \textbf{8.94°}& \textbf{2.35} & \textbf{5.05°}& \textbf{2.33} & \textbf{6.37°}\\
 \hline
\end{tabular}
\caption{Comparison of temporal regularisation methods on camera pose estimations provided by PoseNet on the \emph{Cambridge Landmarks} dataset. As in \cite{Kendall2015}, values are given as median errors in translation [m] and rotation [degrees].} 
\label{tbl:cambridge}
\end{table*}

\begin{figure}[t]
\centering
\includegraphics[width=0.91\columnwidth]{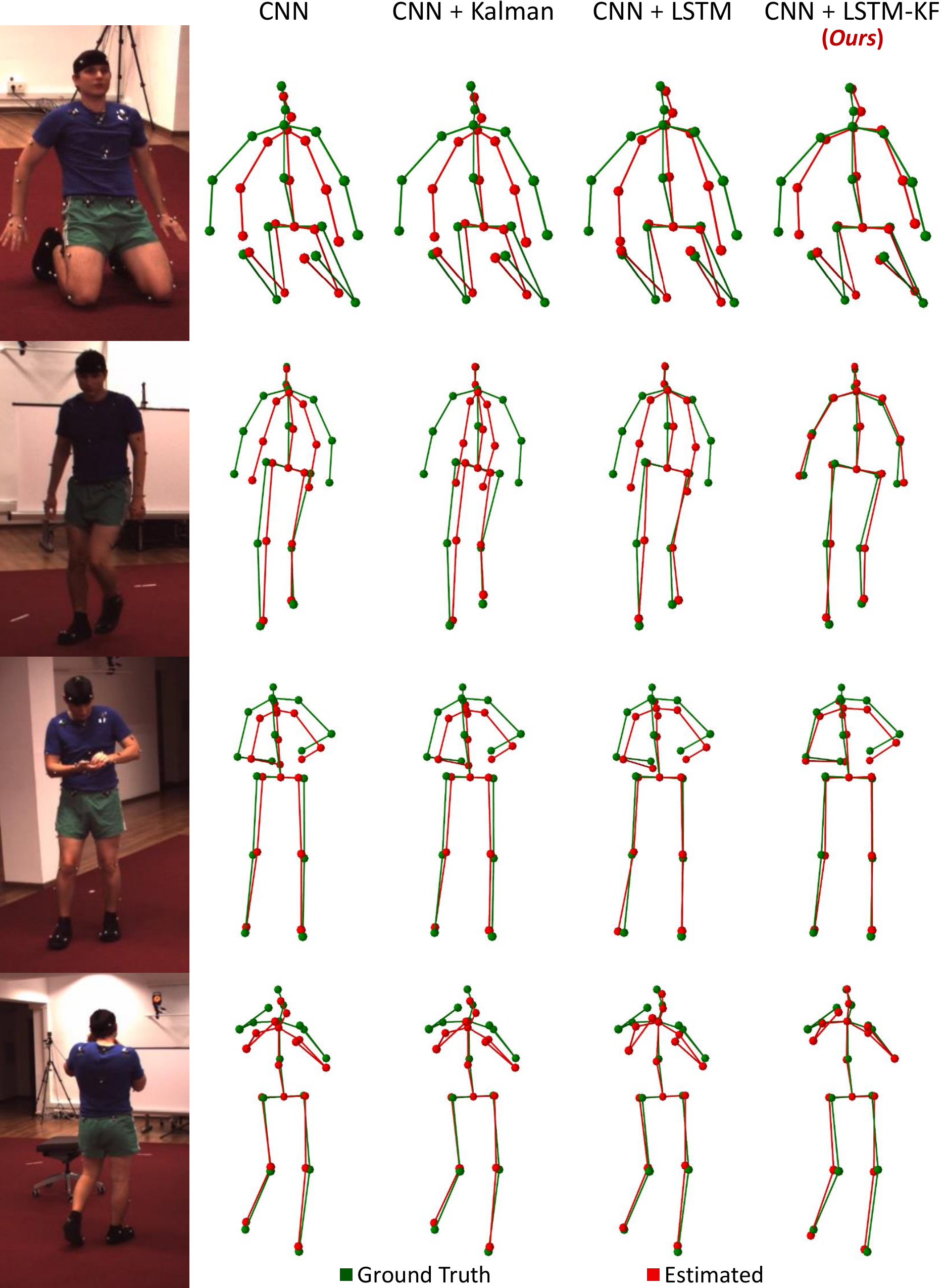}
\caption{Qualitative results on the \textit{Human3.6M} dataset. Ground truth pose in green and estimation in red. Based on the initial CNN estimation, we compare temporal regularization output of Kalman, standard LSTM and our LSTM-KF method. Especially for arm and leg joints, our model improves over the other methods.}
\label{fig:human36m}
\end{figure}

In particular, given the abundance of training samples for this dataset, we employ the bigger network architectures presented in Fig. \ref{architecturedetails}. 
Specifically, LSTM$_f$ consists of 3 stacked layers with 1024 hidden units each, followed by three fully connected (FC) layers with 1024, 1024 and 48 hidden units.
%TODO We need to add this explanation, but currently it kills our page limit.
%The number of layers and number of hidden units was found in smaller experiments on a portion of the full dataset.
The standard LSTM is constructed in the same way as LSTM$_f$. 
We apply the ReLU non-linearity to all FC layer activations except for the last layer, and each LSTM layer is followed by a dropout layer with a keep probability of 0.7. 
LSTM$_Q$ and LSTM$_R$ follow a single layer architecture with 256 hidden units, followed by an FC layer with 48 hidden units.
LSTM-KF and the standard LSTM are trained with a learning rate of 1e-5, with a decay of 0.95 starting from the second epoch. For this training we use truncated backpropagation through time, propagating gradients for 100 time steps.
Qualitative pose estimation results are shown in Figs.~\ref{fig:teaser} and~\ref{fig:human36m} and quantitative pose estimation errors in Table~\ref{tbl:human36m} together with those of four recently published state-of-the-art approaches.
We furthermore show how the estimated measurement noise covariance develops over the course of a test sequence in Fig.~\ref{co_image}.

The results show that the LSTM-KF significantly improves on the raw measurements and outperforms standard LSTM across all actions, achieving on average 14\% improvement over the best state-of-the-art approach. 
Furthermore, as expected, temporal information consistently improves over the raw one-shot estimations from the Inception-v4 model.
It is also relevant to note that the use of the inception architecture alone outperforms previous work.

\begin{table*}[t]
\centering
\begin{tabular}{l c c c c c c c c c c} 
\hline
 & \multicolumn{2}{c}{Kinect Box}  & \multicolumn{2}{c}{Tide} & \multicolumn{2}{c}{Orange Juice}& \multicolumn{2}{c}{Milk}& \multicolumn{2}{c}{\bf{Mean}}\\
  & \emph{tran.} & \emph{rot.} & \emph{tran.} & \emph{rot.} &\emph{tran.} & \emph{rot.} &\emph{tran.} & \emph{rot.} &\emph{tran.} & \emph{rot.}\\
 \hline
   Tan \etal \cite{Tan2015} &  1.70& 	0.30°& 1.17 & 0.44° & 1.29 & 0.35° & 1.27 & 0.41°& 1.36 & 0.37°\\
   + {\small Kalman Vel.} al. &  1.69& 0.29°& 1.84 & \bf{0.38°} & 1.27 & 0.35° & 1.27 & \bf{0.35°} & 1.52 & 0.34°\\      
   + {\small Kalman Acc.} &  1.69& 	\bf{0.28°} & 1.84 & \bf{0.38°} & 1.28 & 0.31° & 1.79 & 0.42°& 1.65 & 0.35°\\
   + {\small EMA} &  1.71& 			\bf{0.28°} & 1.17 & 0.39° & 1.50 & \bf{0.28°} & 1.49 & 0.37°& 1.47 & \bf{0.33°}\\
   + {\small Std. LSTM} &  41.03& 		6.30°& 32.23 & 8.31° & 30.16 & 7.42° & 18.3 & 7.95°& 30.43 & 7.49°\\
   + {\small LSTM-KF (\bf{ours})} & \bf{0.86}& 0.35°& \bf{0.77} & 0.49° & \bf{0.59} & 0.37° & \bf{0.66} & 0.43°& \bf{0.72} & 0.41°\\
\hline
  \end{tabular}
 \caption{We show the effect of temporal regularisation on object tracking estimations of Tan~\etal. We denoting the errors in translation as [mm] and rotation in [degrees]} \label{tbl:tan}
 \end{table*}

\subsection{Camera Tracking} \label{sec:cameratracking}
To demonstrate the wide applicability of our method, we selected camera pose estimation as another application domain and evaluate on the \textit{Cambridge Landmarks}\cite{Kendall2015} and \textit{7 Scenes}\cite{shotton2013scene} datasets. 
The \textit{Cambridge Landmarks} dataset contains 5 different large outdoor scenes of landmarks in the city of Cambridge.
The \textit{7 Scenes} dataset contains 7 image series captured in typical everyday indoor scenes.
Both datasets come with a predefined training and test split that we follow.
In order to generate one-shot camera pose estimates on which we compare the temporal regularisation methods, we retrain the publically avaliable PoseNet CNN architecture \cite{Kendall2015} on the respective training partition of each dataset.

Since these datasets are much smaller than the previously used Human3.6M dataset, we employ the smaller network architectures presented in Fig. \ref{architecturedetails} so to prevent overfitting. 
Specifically, for LSTM$_f$, LSTM$_Q$, and LSTM$_R$ we use a single layer architecture with 16 hidden units, where each LSTM layer is followed by a fully connected layer without non-linearity.
The standard LSTM follows the LSTM$_f$ architecture.
We use batch size of 2, set the learning rate to 5e-4, and train for 10 epochs. Here, we use truncated backpropagation through time, propagating gradients for 10 time steps.

Table \ref{tbl:cambridge} for \emph{Cambridge Landmarks} and Table \ref{tbl:sevenscenes} for \emph{7 Scenes} show the quantitative results on those datasets.
Our approach consistently improves estimations on the \emph{7 Scenes} dataset.
The same is true for the \emph{Cambridge Landmarks} dataset, except for the \emph{King’s College} and \textit{S. Facade} sequence.
In the \emph{King’s College} sequence, learning the motion model might be a disadvantage, as the camera trajectory in the training set moves in curves, while in the test set it resembles a straight line.
The \textit{S. Facade} sequence poses a different challenge for the LSTM-KF, as its training set only consists of 231 frames, which is most likely too short for the LSTM$_f$ to learn a valid motion model (average training sequence length: 1370 frames).
Since the datasets are quite limited in size, the standard LSTM was not able to improve the results, and even decreases the accuracy.
Our LSTM-KF model achieves an improvement of up to 6.23\% for translation and 7.53\% for rotation on average over the \emph{Cambridge Landmarks} dataset, while \textit{Kalman Vel} and \textit{Kalman Acc} improve 4.1\% and 4.43\% for translation and 1.66\% and 1.17\% for rotation, respectively.
For the \emph{7 Scenes} dataset, LSTM-KF improves the PoseNet estimations about 10.13\% for translation and 7.53\% for rotation.
\textit{Kalman Acc}, \textit{Kalman Vel} and standard LSTM algorithms were not able to improve over the original PoseNet estimation.

\subsection{Object Tracking}
As third experiment, we evaluated our method on the public \textit{MIT RGB-D Object Pose Tracking Dataset}~\cite{Choi2013}.
As in Tan~\etal~\cite{Tan2015}, we used four synthetically generated object tracking sequences from the dataset, for which 6-DOF ground truth poses were available.
The sequences consist of 1,000 RGB-D frames in which the tracked object (\textit{Kinect Box, Milk, Orange Juice, Tide}) was rendered in front of a virtual kitchen scene.

Our model parameters were set up equal to experiment~\ref{sec:cameratracking}, specifically using single layer LSTMs with 16 hidden units, a batch size of 2 and a learning rate of 5e-4. We trained for 120 epochs, again using truncated backpropagation through time, propagating gradients for 10 time steps.
The same holds true for the standard LSTM method that we evaluated against.
As no separate training set was provided, we performed 2-fold cross validation by training on the \textit{Kinect Box} and \textit{Milk} sequence to test on \textit{Orange Juice, Tide} and vice versa.
As input to all methods, we use the raw object pose estimations of~\cite{Tan2015}, which were provided by the authors.
This tracking algorithm exploits successive frame pairs to estimate the 3D pose of a 3D CAD model being tracked through a sequence of depth frames.
Hence, the task for all methods compared in this experiment is to gain additional improvements over an existing object tracking method.
Results for this scenario are reported in Table \ref{tbl:tan}. 
The methods that did not learn the motion model on training data, i.e.\ \textit{Kalman Vel, Kalman Acc} and \textit{EMA}, were not able to meaningfully improve on the translation estimation, while rotation was slightly improved.
For the object position, LSTM-KF achieves the best results at 0.72\,mm average error, improving 47.05\,\% over the original estimation.
The standard LSTM approach yields a high error in both position and rotation estimation.
It does not follow the measurement and starts to deviate from the correct trajectory rather quickly.
We assume that the task of implicit fusion of past state and measurement update is too difficult for the standard LSTM to learn, given the available training data.

\section{Conclusions}
In this work, we introduced the long short-term memory Kalman filter (LSTM-KF). This model alleviates the modeler from specifying motion and noise models a priori and simultaneously allows the learning of rich models from data which are extremely difficult to write down explicitly. In an extensive set of experiments, we found that the LSTM-KF outperforms both the standalone Kalman filter and standalone LSTM for temporal regularization. In addition, we achieved state-of-the-art performance on three diverse tasks, for example reducing the joint error in the Human 3.6M dataset by 13.8\%, from 82.3\,mm to 71.0\,mm.

\section{Acknowledgments}
The authors would like to thank David J. Tan for the fruitful discussions and support in preparation of this work. 

{\small
\bibliographystyle{ieee}
\bibliography{ICCV_KLSTM_bibliography}
}

\end{document}